\documentclass[letterpaper, 10 pt, conference]{ieeeconf}

\usepackage[utf8]{inputenc}
\usepackage[T1]{fontenc}
\usepackage{acronym}

\usepackage{array}
\usepackage{float}
\usepackage{amsfonts}
\usepackage{amsmath}
\usepackage{amssymb}
\usepackage{mathtools}

\DeclareMathOperator*{\argmin}{\arg\!\min}
\DeclareMathOperator*{\argmax}{\arg\!\max}
\DeclarePairedDelimiterX{\infdivx}[2]{(}{)}{%
	#1\;\delimsize\|\;#2%
}
\newcommand{\infdiv}{D\infdivx}

\usepackage[english]{babel}

\usepackage{wrapfig}
\usepackage[export]{adjustbox}
\usepackage{microtype}
\usepackage[dvipsnames]{xcolor}
\usepackage{graphicx}
\usepackage{graphics}
\usepackage{svg}
\usepackage{tikz}

\usepackage{multicol}
\usepackage{multirow}
\usepackage[font=footnotesize]{caption}
\usepackage{subcaption}
\usepackage{booktabs}

\usepackage{authblk}
\usepackage{algorithm, algorithmicx}
\usepackage[noend]{algpseudocode}

\usepackage{todonotes}
\usepackage{listings}

\renewcommand{\thetable}{\arabic{table}}

\graphicspath{{/images}}
\makeatletter
\def\endthebibliography{%
	\def\@noitemerr{\@latex@warning{Empty `thebibliography' environment}}%
	\endlist}
\makeatother

\definecolor{label-running} {RGB}{ 31,119,180}
\definecolor{label-walking} {RGB}{255,127, 14}
\definecolor{label-jumping} {RGB}{ 44,160, 44}
\definecolor{label-standing}{RGB}{148,103,189}
\definecolor{label-sitting} {RGB}{140, 86, 75}
\definecolor{label-lying}   {RGB}{127,127,127}
\definecolor{label-falling} {RGB}{188,189, 34}
\definecolor{label-transit} {RGB}{ 23,190,207}

\IEEEoverridecommandlockouts

\title{\LARGE \bf Task and Domain Adaptive Reinforcement Learning for Robot Control}
\author{Yu Tang Liu$^{1,2}$, Nilaksh Singh$^{3,2}$, and  Aamir Ahmad$^{2,1}$	\thanks{$^1$Max Planck Institute for Intelligent Systems, 72076 T{\"u}bingen, Germany. $^2$University of Stuttgart, 70569 Stuttgart, Germany. $^3$IIT Kharagpur, West Bengal, India, 721302. \texttt{yutang.liu@tuebingen.mpg.de, nilaksh404@kgpian.iitkgp.ac.in, aamir.ahmad@ifr.uni-stuttgart.de}}}

\makeatletter
\let\NAT@parse\undefined
\makeatother

\usepackage[english]{babel}
\usepackage{hyperref}

\begin{document}

	\maketitle 

	\makeatletter

\makeatother
\begin{abstract}
Deep reinforcement learning (DRL) has shown remarkable success in simulation domains, yet its application in designing robot controllers remains limited, due to its single-task orientation and insufficient adaptability to environmental changes.
To overcome these limitations, we present a novel adaptive agent that leverages transfer learning techniques to dynamically adapt policy in response to different tasks and environmental conditions. 
The approach is validated through the blimp control challenge, where multitasking capabilities and environmental adaptability are essential. 
The agent is trained using a custom, highly parallelized simulator built on IsaacGym. We perform zero-shot transfer to fly the blimp in the real world to solve various tasks. We share our code at \url{https://github.com/robot-perception-group/adaptive\_agent/}.
\end{abstract}



\section{Introduction}
\label{sec:Introduction}
One promising approach to achieve optimal control in robotics is through reinforcement learning (RL) \cite{bertsekas1996neuro}, an algorithm for searching control policies by environmental interaction.
Despite its potential, RL methods remain impractical for most robotics applications, primarily due to the agents' inability to adapt to changes in key components of the Markov decision process (MDPs), such as reward function and transition dynamics.

In this work, we developed an adaptive agent that is robust against those changes or capable of adapting to those changes, by leveraging transfer learning (TL) techniques. Specifically, we focus on task transfer \cite{zhu2023transfer} and domain transfer \cite{zhao2020sim}. Task transfer enhances the agent's ability to apply previously acquired skills to new tasks, thereby allowing building complex skills from simpler ones. Domain transfer, on the other hand, facilitates adaptation to varying transition dynamics, especially crucial for bridging the sim-to-real gap.

\begin{figure}[t!]
	\centering
	\includegraphics[width=0.5\textwidth]{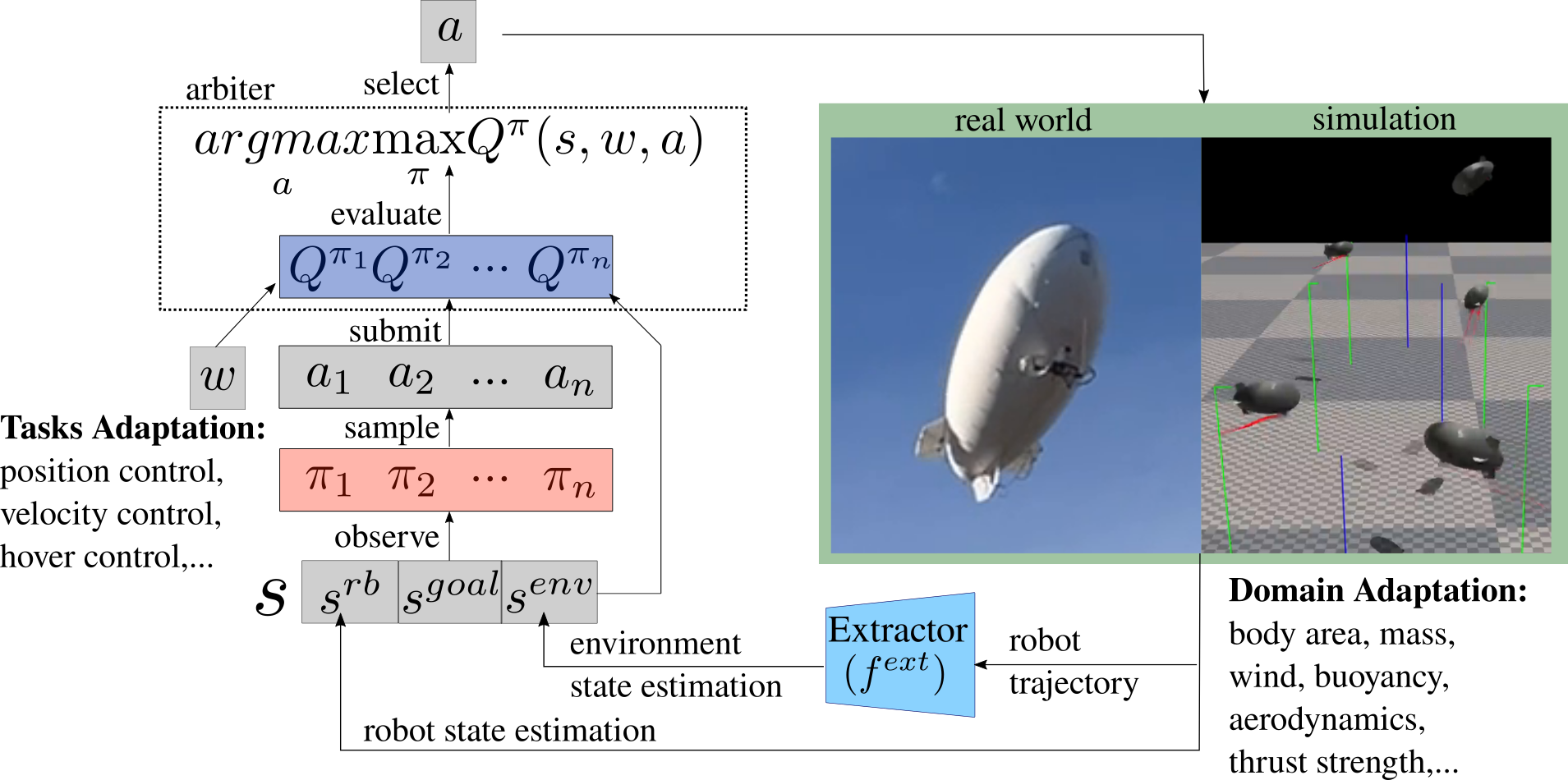}	
	\caption{The arbiter architecture allows our adaptive agent trained purely in simulation to achieve zero-shot domain transfer and sim-to-real transfer, to control the real robot and perform unseen tasks. The arbiter selects action by observing the current task weight $w$, state of the robot $s^{rb}$, goal $s^{goal}$ and environment $s^{env}$. Our customized blimp simulation based on IsaacGym can support a high degree of parallelization to facilitate multitask learning. Each environment possesses a green and a blue pole corresponding to the hover and navigation goals' positions, respectively. The agent should navigate the blimp to different goal areas depending on the task specification.}
	\label{fig: illustration}
\end{figure}

We introduce two fundamental modules developing our adaptive agent: (1) an architecture, which we call Arbiter-SF, that facilitates task transfer, and (2) a robust feature extractor. 
The task transfer module is inspired by the arbiter architecture \cite{russell2003q}, which involves an arbiter and multiple sub-policies, or "primitives," each specialized in handling specific sub-tasks. At every time step, these primitives propose their actions to the arbiter, which then evaluates these suggestions before the action selection. 
To enable task and domain transfer, we extend the arbiter observations with task specifications and the environment states (Fig.~\ref{fig: illustration}). To enhance task transfer performance, we integrate a \textit{successor feature} (SFs, \cite{barreto2020fast}) into the arbiter framework, which allows specifying a task as a continuous vector and enables real-time task transfer. 
Secondly, to adapt to varying dynamics, we infer environment states via a feature extractor by reading a snapshot of past interactions. To train this extractor, our approach adopts a robust two-stage training procedure, Rapid Motor Adaptation (RMA) \cite{kumar2021rma}, which trains an encoder first to capture the environmental factors within a domain randomized simulation and then replicates the representation by the extractor. 

\begin{figure*}[t!]
	\centering
	\includegraphics[width=0.87\textwidth]{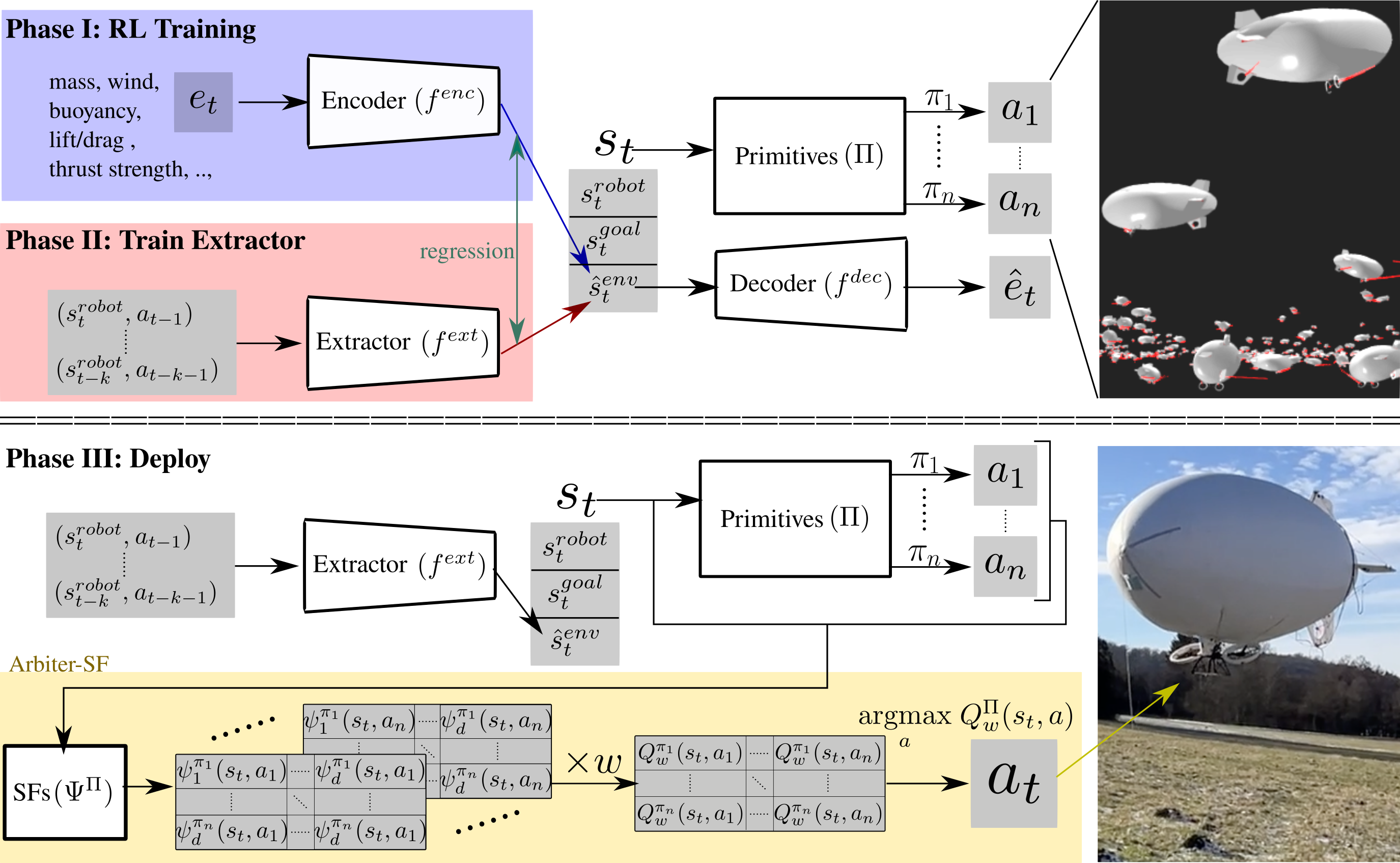}	
	\caption{\textbf{Our proposed methodology}: the adaptive agent follows the RMA training procedure. In the first phase, all modules except the feature extractor undergo training, focusing on training the primitives and constructing informative environmental states.  
		In the second phase, the extractor is trained to replicate the encoder's output, ensuring that it captures the environment state. A decoder is utilized in both phases to refine the environmental representation. During training, each primitive may control several blimps simultaneously.
		The arbiter is introduced during the deployment phase for selecting an action.}
	\label{fig: arbiter_architect}
\end{figure*}

To validate our approach, we tested the adaptive agent in the autonomous blimp control challenge. Given a blimp's extended hover ability, they will undertake various control tasks, from basic maneuvers like hover, navigation, and take-off to activities such as aerial target tracking \cite{price2023driven}. By learning these tasks together, which share a common state space, we can potentially accelerate learning and enhance the sample efficiency of the RL agent.
Nevertheless, blimps' advantages come at the cost of complex dynamics: a highly nonlinear, time-delayed, environment-dependent system requiring policies that are either robust or adaptable to varying air conditions such as temperature, wind, and buoyancy. 
To address the computational demands of ROS/Gazebo simulation used in previous works \cite{liu2022deep}, we developed a blimp simulator in IsaacGym \cite{liang2018gpu}, enabling significantly higher parallelization that facilitates multi-task training.

We summarize the contributions as follows:
\begin{itemize}	
	\item A novel adaptive RL agent via task and domain transfer that is tailored to the robotics applications. This is achieved through an Arbiter-SF architecture, a feature extractor, and an RMA training procedure.
	\item A customized blimp simulation, based on IsaacGym \cite{liang2018gpu}, supporting thousands of parallel simulations.
\end{itemize}
 

\section{Related work}
\label{sec:2_related}

A direct approach to obtain a \textbf{multi-task agent} is augmenting task representations to the observation \cite{espeholt2018impala}, which could demand extensive training data.
An alternative is task transfer, where a trained policy is recycled to solve a new task such that agents can progressively handle harder challenges and learn a repertoire of skills \cite{kalashnikov2021mt}. 
There exist numerous task transfer methods.
Primitive-based methods train a gating network for primitives composition \cite{peng2019mcp}, or chain in temporal sequences, e.g. hierarchical RL \cite{pateria2021hierarchical}, options \cite{sutton1999between}, etc. 
Modularized methods obtain a single multi-task policy via various sub-networks \cite{yang2020multi}, or policy distilling \cite{ghosh2017divide}. 
Alternatively, representation learning develops a shared representation \cite{gupta2017learning} or an explicit dynamics model \cite{zhang2018decoupling} for solving new tasks. 
In contrast, our work adopts a unique kind of value function, i.e. SFs \cite{dayan1993improving}, facilitating \textit{zero-shot} task transfer \textit{without} augmenting task representation \cite{liu2024multi}. 
Our adaptive agent, built on top of the arbiter architecture \cite{russell2003q} and SFs' policy improvement theory \cite{barreto2020fast}, targets the \textit{continuous control} domain and uses value functions for primitive composition, \textit{without} training an additional gating network.

To bridge the \textbf{domain gaps} \cite{zhao2020sim}, domain randomization techniques enhance policy robustness by training across varied simulations \cite{tobin2017domain}, while domain adaptation aligns representations across source and target domains \cite{taylor2009transfer}, often employing imitation learning \cite{ho2016generative}. Few-shot learning adapts to a new domain, either using gradient update \cite{wang2016learning}, or estimate environment state \cite{peng2020learning}. Our method is based on the RMA zero-shot approach to obtain a feature extractor that estimates environment state \cite{kumar2021rma}.


Prior work on \textbf{blimp control} uses ROS/Gazebo simulation and cascade PID controllers \cite{price2020simulation}, encountering difficulties due to the blimps' intricate dynamics.  
To handle environmental uncertainties like wind, buoyancy, and weight distribution, robust and adaptive control methods are desired. Learning-based approaches can potentially address these challenges \cite{rottmann2009adaptive}. 
A deep residual RL framework for position control has demonstrated superior performance by integrating classic methods \cite{liu2022deep}. In this work, we instead focus on enhancing multitasking capabilities and domain adaptability.

%

\setlength{\belowdisplayskip}{0pt} \setlength{\belowdisplayshortskip}{0pt}
\setlength{\abovedisplayskip}{0pt} \setlength{\abovedisplayshortskip}{0pt}

\section{Methodology}
\label{sec:Methodology}

\subsection{Problem Definition}
The goal of an adaptive agent is to search a set of domain adaptive primitives for task transfer via an arbiter. An adaptive agent improves a primitive through interaction with the environment. It formulates this procedure by MDPs $\mathcal{M}_w\equiv<\mathcal{S},\mathcal{A}, \mathcal{T},R_w,\gamma>$, where $\mathcal{S}$ and $\mathcal{A}$, define the state and action space respectively. 
The dynamics $\mathcal{T}(s_{t+1}|s_t,a_t): \mathcal{S}\times\mathcal{A}
\rightarrow\mathcal{S}$, defines the transition probability from one state to another and $\gamma\in[0,1]$ is the discount factor. 
Each of the MDPs has a unique reward function $r_t=R_w(s_t, a_t, s_{t+1}): \mathcal{S}\times\mathcal{A}
\times\mathcal{S}\rightarrow \mathbb{R}$ specifies the desired task. 
We assume a linear reward function that can be decomposed as follows, $R_w(s_t, a_t, s_{t+1})=\mathbf{\Phi}(s_t, a_t, s_{t+1})^\top \cdot w$, where $w\in \mathbb{R}^d$ is the task weight and $\boldsymbol{\phi}_t=\mathbf{\Phi}(s_t, a_t, s_{t+1}) \in \mathbb{R}^d$ is the features, which are relevant to task solving. 

In task transfer, we consider a set of MDPs $\{\mathcal{M}_w |w \in \mathcal{W}\}$, and each member MDP possesses a unique task weight defined in the task space $\mathcal{W}$ with the features shared with all other member MDPs. 
The goal is to find a set of $n$ primitives, $\Pi=\{\pi_{w_1}, ...,\pi_{w_n}\}$, where each primitive, $ \pi_{w_i}(a|s): \mathcal{S} \rightarrow \mathcal{A}$, is trained to solve its primitive task, $w_i \in \mathcal{W}' \subset \mathcal{W}$, by maximizing the expected discounted return, $J(\pi_{w_i})=\mathbb{E}_{\pi_{w_i}, \mathcal{M}_{w_i}}\left[\sum_{t=0}^{\infty} \gamma^{t}r_{w_i,t}\right]$, such that we can recover the policies $\pi_w$ of the unseen tasks $w \in \mathcal{W}$, by composing these primitives that only solves a subset of tasks $\mathcal{W}'$. 

One way to transfer between domains is to extract the environment state from the observation. We model robot state, goal, together with the environment state, as part of the state space, i.e. $S=(S^{robot}, S^{goal}, S^{env})$, where $s^{robot}_t \in S^{robot}$ can be observed via sensors or from a state estimator, $s^{goal}_t\in S^{goal}$ is virtual goal states specified by user, and $s^{env}_t\in S^{env}$ is the environment state un- (or partially) observable by the sensors. 
We aim to find a feature extractor, $f^{ext}$ that can extract environment state $s_t^{env}$ from a trajectory snapshot $\tau_{t:t-k} = (s^{robot}_t, a_{t-1},..., s^{robot}_{t-k}, a_{t-k-1})$, i.e. $\hat{s}^{env}_t=f^{ext}(\tau_{t:t-k})$. Then the full state can be represented as $s_t=(s_t^{robot}, s_t^{goal}, \hat{s}_t^{env})$. 

In summary, the task transfer problem boils down to training primitives $\Pi=\{\pi_{w_1}, ...,\pi_{w_n}\}$ for solving tasks $\{w_1,...,w_n\}$ such that an arbiter can compose them for solving any tasks $w$. The domain transfer problem becomes training a feature extractor $f^{ext}(\tau_{t:k})$ to reveal the environment state $s^{env}_t$.

\subsection{Task Transfer via Successor Feature-based Arbiter}
When training the primitives, a set of value functions, or Q-functions, $\boldsymbol{Q}^{\Pi}=\{Q^{\pi_{w_1}},...,Q^{\pi_{w_n}}\}$ are introduced. A Q-function is defined as:
\begin{align}
	Q^{\pi_{w_i}}_{\mathcal{M}_{w_i}}(s_t, a_t) \equiv \mathbb{E}_{\substack{\mathcal{M}_{w_i}, \pi_{w_i}}}\left[ \sum^{\infty}_{k=0}\gamma^k r_{w_i, \scriptstyle{t+k}}| s_t,a_t \right], \label{eqn: action-value function}
\end{align}
which estimates the total discounted reward of task $w_i$ by following primitive $\pi_{w_i}$. 
One can see a value function as a performance report of a primitive, i.e., the higher the value indicating a better performance of the primitive in solving a task. Therefore, given a set of primitives, the arbiter is an operator, $argmax_{\pi_w \in \Pi}Q^{\pi_w}$, reading values of all primitives for selecting the best one to perform the action. For brevity, we define $Q^{\pi_{w_i}}_{w} \equiv Q^{\pi_{w_i}}_{\mathcal{M}_{w}} $, meaning the performance of primitive $\pi_{w_i}$ in solving task $w$.
Given that  $R_{w_i}(s_t, a_t, s_{t+1})=\boldsymbol{\Phi}(s_t, a_t, s_{t+1})^\top \cdot w_i$, we can rewrite the Q-function in following form, 
\begin{align}
	&Q^{\pi_{w_i}}_{w_i}(s_t, a_t)=\mathbb{E}_{\mathcal{M}_{w_i}, \pi_{w_i}}\left[ \sum^{\infty}_{k=0}\gamma^i r_{w_i,  \scriptstyle{t+k}} \right] \\
	&=\mathbb{E}_{\mathcal{M}_{w_i}, \pi_{w_i}}\left[\sum_{k=0}^{\infty} \gamma^{k}\boldsymbol{\phi}^\top_{t+k} \right]\cdot w_i 
	=\boldsymbol{\psi}^{\pi_{w_i}}(s_t,a_t)^\top \cdot w_i \label{eqn:sf.w}
\end{align}
where $\boldsymbol{\psi}^{\pi_{w_i}}(\cdot)\in \mathbb{R}^d$ is the \textit{successor feature} (SFs) of the primitive $\pi_{w_i}$. 
Since Q-function can be viewed as a performance metric of a primitive,  SFs provide a shortcut to estimate primitive performance in different tasks $w$, i.e. $Q^{\pi_{w_i}}_{w}(s_t,a_t)=\boldsymbol{\psi}^{\pi_{w_i}}(s_t,a_t)^\top \cdot w$. 

In the SF-based arbiter framework, training a number of primitives $\Pi$ induces their associated SFs instead of Q-function, i.e., $\Psi^{\Pi}=\{\boldsymbol{\psi}^{\pi_{w_1}},...,\boldsymbol{\psi}^{\pi_{w_n}}  \}$. Thus, depending on the current task $w$, the arbiter can select primitive actions based on profiles provided by the SFs  \cite{barreto2020fast}, i.e. 
\begin{align}
	\begin{split}
		\argmax_{a_t \sim \pi } &\max_{Q^{\pi}_w \in Q^{\Pi}} Q^{\pi}_w(s_t,a_t) \equiv\\
		&\argmax_{\boldsymbol{a}_t \sim \Pi } \boldsymbol{Q}^{\Pi}_w(s_t,\boldsymbol{a}_t) =  \argmax_{\boldsymbol{a}_t \sim \Pi } \boldsymbol{\Psi}^{\Pi}(s_t,\boldsymbol{a}_t)^\top \cdot w. \label{eqn: gpi}
	\end{split}
\end{align}
which means to sample actions from each primitive and select the action incurring the largest primitive Q-values estimated via primitve SFs. Fig.~\ref{fig: arbiter_architect} summarizes the action selection procedure.

\subsection{Successor Feature Policy Evaluation and Improvement}
We can obtain primitives and their SFs via soft SFs policy iteration \cite{liu2024multi}, which iterates between a soft SF policy evaluation step and a soft policy improvement step. 
The evaluation step improves the SFs estimation as follows,
\begin{align}
	\begin{split}
		\boldsymbol{\psi}^{\pi_w}&(s_t,a_t) \\
		&= \boldsymbol{\phi}_t(s_t,a_t) + \gamma \mathbb{E}_{s_{t+1}\sim p} \left[ \alpha \log \int_{a \sim \pi_w}e^{\boldsymbol{\psi}^{\pi_w}(s_{t+1},a)/\alpha}) \right], \label{eqn:sf_evaluation}
	\end{split}
\end{align}
where $\alpha \in \mathbb{R}^d$ is a temperature parameter for controlling the exploration of each primitive. Then the soft policy improvement step aims to replicate policy distribution based on the Q-values of each action, i.e., the larger the Q-value, the higher the probability of sampling the action, $\pi_w(\cdot|s) \propto e^{Q^{\pi_w}_w(s, \cdot)}$, which is achieved by solving,
\begin{align}
	\pi_w(\cdot|s)=\argmin_{\pi_w'}\infdiv*{\pi_w'(\cdot|s_t)}{\frac{e^{Q^{\pi_w}_{w}(s_t,\cdot)}}{Z(s_t)}}, \label{eqn: soft successor feature policy improvement}
\end{align}
where $Q^{\pi_w}_w$ is derived by (\ref{eqn:sf.w}), \textit{D}$(\cdot||\cdot)$ measures the KL divergence, and $Z(s_t)=\int_{\mathcal{A}} e^{Q^{\pi_w}_w(s_t,\cdot)}$ is a negligible function as it does not contribute to gradient computation. Alternating between an evaluation step and an improvement step guarantees convergence to the optimal primitive $\pi_{w_i}^*$ and the optimal Q-function $Q^{*,\pi_{w_i}}\equiv Q^{\pi_{w_i}^*}$ yielded from the optimal SFs $\boldsymbol{\psi}^{*,\pi_{w_i}}\equiv \boldsymbol{\psi}^{\pi_{w_i}^*}$ of each primitive.

\subsection{Training Scheme}
The RMA \cite{kumar2021rma} unfolds training in three phases (Fig.~\ref{fig: arbiter_architect}): Phase I: Training the primitive module $\Pi$, SFs module $\Psi^{\Pi}$ and environment encoder $f^{enc}$. Phase II: Replicating the encoder $f^{enc}$ representation using the feature extractor $f^{ext}$. Phase III: Evaluation/deployment phase.

\subsubsection{Phase I: Training the SFs and Primitive Network}
The primitives are trained by alternating between an interaction phase and a training phase. During the interaction phase, training data is gathered by interaction within the environment in parallel and stored in a replay buffer. In the training phase, we sample data from the replay buffer and update the SFs and primitives via (\ref{eqn:sf_evaluation}) and (\ref{eqn: soft successor feature policy improvement}), respectively. 

Every training episode, the simulation exposes randomized environment factors of length $E$, i.e., $e \in \mathbb{R}^E$. Then the encoder, parameterized by $\theta_{f^{enc}}$, learns to convert these factors to the environment state, $\hat{s}_t^{env}=f^{enc}(e_t)$. Then the state variable is a concatenation of the robot state, goal state, and the environment state $s_t=(s_t^{robot}, s_t^{goal}, \hat{s}_t^{env})$, where robot states and goal states are randomized for each training episode.

In total, we have $n$ pairs of SFs and primitives with the $i$-th successor feature $\boldsymbol{\psi}^{\pi_{w_i}}$, target network $\bar{\boldsymbol{\psi}}^{\pi_{w_i}}$, and the primitive network $\pi_{w_i}$ parameterized by $\theta_{\boldsymbol{\psi}^{\pi_{w_i}}}$, $\theta_{\bar{\boldsymbol{\psi}}^{\pi_{w_i}}}$, and $\theta_{\pi_{w_i}}$ respectively. The target network is a common practice in RL to reduce the overestimation of the values. The loss for the i-th successor feature is computed as follows,
\begin{align}
	J_{\psi^{\pi_{w_i}}}(\theta_{\psi^{\pi_{w_i}}}) &= \mathbb{E}_{(s_t,a_t)\sim\mathcal{D}}    \left[     || \boldsymbol{\psi}^{\pi_{w_i}}(s_t,a_t) - \hat{\boldsymbol{\psi}}^{\pi_{w_i}}(s_t,a_t)  ||_2	\right], \label{eqn: sfloss}
\end{align}
where $\mathcal{D}$ is the replay buffer and $||\cdot||_2$ denotes L2 norm. We have the TD-target $\hat{\boldsymbol{\psi}}^{\pi_{w_i}}$ approximated by the target network $\bar{\psi}^{\pi_{w_i}}$,
\begin{align}
	\hat{\boldsymbol{\psi}}^{\pi_{w_i}}(s_t,a_t)= \boldsymbol{\phi}_t + \gamma \mathbb{E}_{\substack{s_{t+1}\sim \mathcal{D} \\ a_{t+1} \sim \pi_{w_i}}}\left[  \boldsymbol{\bar{\psi}}^{\pi_{w_i}}(s_{t+1},a_{t+1}) \right].
\end{align}
Then the primitive can be improved by minimizing the objective,
\begin{align}
	J_{\pi_{w_i}}(\theta_{\pi_{w_i}})=-\mathbb{E}_{\substack{s_t\sim \mathcal{D} \\ a_t \sim \pi_{w_i}}} \left[ Q^{\pi_{w_i}}_{w_i}(s_t, a_t) - \alpha \log \pi_{w_i}( a_t|s_t )   \right], \label{eqn: primitive loss}
\end{align}
where $Q^{\pi_{w_i}}_{w_i}(s_t, a_t)=\boldsymbol{\psi}^{\pi_{w_i}}(s_t,a_t)^\top \cdot w_i$ and $-\log\pi(\cdot|s)$ is an estimator for the policy entropy. 

Note that, in theory, we can update the encoder $\theta_{f^{enc}}$ via policy and SFs loss. However, in practice, we found training much more stable if only the SFs loss is applied. 

\subsubsection{Phase II: Training the Feature Extractor}
We train a feature extractor parameterized by $\theta_{f^{ext}}$ to replicate the environment state predicted by the encoder by minimizing the loss, 
\begin{align}
	J_{f^{ext}}(\theta_{f^{ext}})=  ||f^{enc}(e_t)-f^{ext}(\tau_{t:t-k})||_2. \label{eqn: extractor loss}
\end{align}
We observed instability in training when initiated with only the feature extractor (Fig.\ref{fig: domain_rand}). Therefore, training an encoder first becomes necessary. 

Lastly, during deployment (Phase III), we use arbiter for composing primitive actions using (\ref{eqn: gpi}). Algorithm~\ref{alg:summary} summarizes how the adaptive agent trains each network module. 
Note that, to keep the computation time constant, in practice, the \textit{for}-loop is replaced with vectorized computations.

\begin{figure*}
	\centering
	\captionsetup[subfigure]{labelformat=empty}
	\begin{subfigure}[t]{0.90\textwidth}
		\centering
		\includegraphics[width=0.99\textwidth]{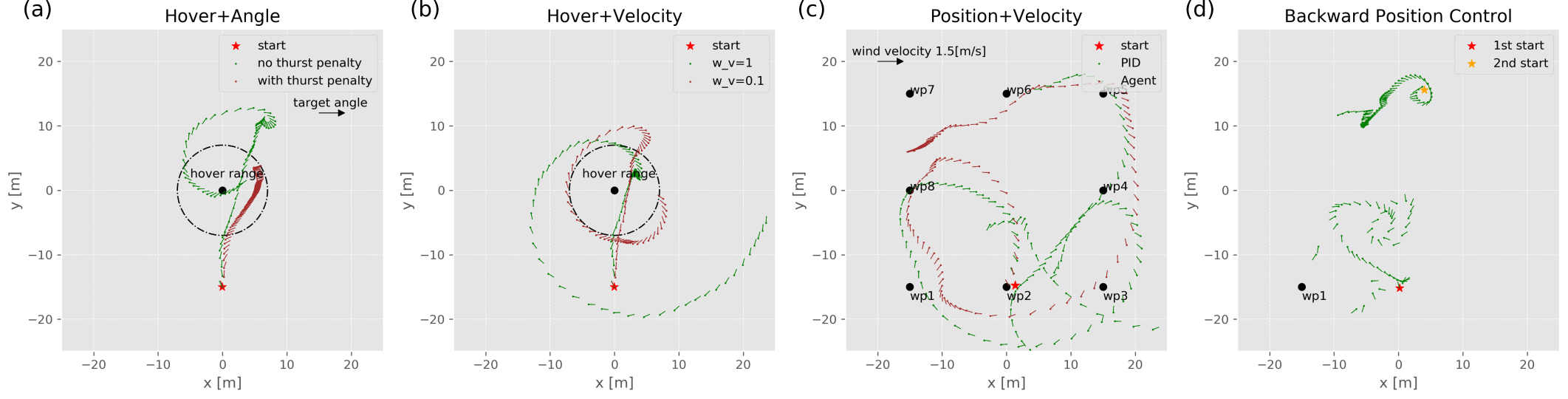}
		\phantomsubcaption
		\label{fig:simtasks_a}
	\end{subfigure}
	\begin{subfigure}[t]{0pt}
		\centering
		\phantomsubcaption
		\label{fig:simtasks_b}
	\end{subfigure}
	\begin{subfigure}[t]{0pt}
		\centering
		\phantomsubcaption
		\label{fig:simtasks_c}
	\end{subfigure}
	\begin{subfigure}[t]{0pt}
		\centering
		\phantomsubcaption
		\label{fig:simtasks_d}
	\end{subfigure}
	\caption{\textbf{Tasks performed in the simulation:} Various kind of tasks can by achieved by mixing components in the task weight. 
		(a) The hover task $w[4]$ can be combined with yaw control task $w[5]$ to track desired angle in the hover zone $w=[0,0,0,0, 1,1,0, 0,0, 0,0]$. We can also add a thrust penalty ($w[10]=0.1$) to prevent overshooting (brown). 
		(b) The hover task $w[4]$ can be combined with velocity control $w[6]$ to track desired velocity in the hover zone $w=[0,0,0,0, 1,0,1, 0,0, 0,0]$. We can reduce the velocity weight to emphasize less on this sub-task (brown), i.e., $w_v=w[6]=0.1$. 
		(c) A higher control performance in waypoint trigger task $w[2]$ can be achieved by mixing position $w[0]$ and velocity control tasks $w[7]$, i.e. $w=[.3,.1,1,0, 0,0,0, .3,0, 0,0]$. Compared with a velocity PID controller (brown), the agent does not blindly follow the velocity commands to prevent overshooting. (d) The agent can fly the blimp backward $w=[0,0,0,-1, 0,0,-1,  0,.1, 0,0]$ by reversing the yaw $w[3]$ and heading velocity task $w[6]$. However, backward flight is slow and inefficient. The agent can hardly maintain its altitude and ends up being reset by the simulator.
		\label{fig:simtasks}}
\end{figure*}

\begin{algorithm}
	\caption{Adaptive RL Agent}\label{alg:summary}
	\begin{algorithmic}[1]
		\footnotesize
		\Statex Initialize network parameters $\theta_{\psi^{\pi}}, \theta_{\bar{\psi}^{\pi}}, \theta_{\pi}, \theta_{f^{enc}}, \theta_{f^{ext}}$
		
		\While{Training} 
		\If{Phase==1} $\hat{s}_t^{env} = f^{enc}(e_t)$   
		\Else $~\hat{s}_t^{env} = f^{ext}(\tau_{t:t-k})$ 				
		\EndIf
		\State $s_t=(s_t^{robot}, s_t^{goal}, \hat{s}_t^{env})$  
		\\
		\If{exploration} $a_t\leftarrow$ Uniform($\mathcal{A}$)  
		\ElsIf{Phase==1 or 2} 
		\State $a_t\leftarrow \pi_i(a|s_t), \quad w_i\sim\mathcal{W}'$
		\ElsIf{Phase==3}
		\State $a_t\leftarrow \argmax_{\boldsymbol{a}_t \sim \Pi }\boldsymbol{\Psi}^{\Pi}(s_t,\boldsymbol{a}_t)^\top \cdot w, \quad w\sim\mathcal{W}$ 				
		\EndIf
		\State $\mathcal{D}\leftarrow \mathcal{D} \cup (s_t,a_t,s_{t+1},\Phi(s_t,a_t,s_{t+1}))$  
		\\
		\If{Phase==1}
		\For{$i \leftarrow 1,2,...,n$}
		\State $\theta_{\psi^{\pi_{w_i}}}\leftarrow \theta_{\psi^{\pi_{w_i}}}- \lambda_{\psi} \nabla_{\theta_{\psi^{\pi_{w_i}}}}J_{\psi^{\pi_{w_i}}}(\theta_{\psi^{\pi_{w_i}}}) $
		\State $\theta_{\pi_{w_i}}\leftarrow \theta_{\pi_{w_i}}- \lambda_{\pi} \nabla_{\theta_{\pi_{w_i}}}J_{\pi_{w_i}}(\theta_{\pi_{w_i}}) $
		\State $\theta_{\bar{\psi}^{\pi_{w_i}}} \leftarrow \tau \theta_{\bar{\psi}^{\pi_{w_i}}} + (1-\tau) \theta_{\psi^{\pi_{w_i}}}$ 
		\EndFor
		
		\ElsIf{Phase==2} \State $\theta_{{f^{ext}}}\leftarrow \theta_{{f^{ext}}}- \lambda_{f} \nabla_{\theta_{f^{ext}}}J(\theta_{f^{ext}}) $				
		
		\ElsIf{Phase==3} pass
		\EndIf
		
		\EndWhile
		
	\end{algorithmic}
\end{algorithm}


\begin{figure*}
	\centering
	\captionsetup[subfigure]{labelformat=empty}
	\begin{subfigure}[t]{0.8\textwidth}
		\centering
		\includegraphics[width=0.99\textwidth]{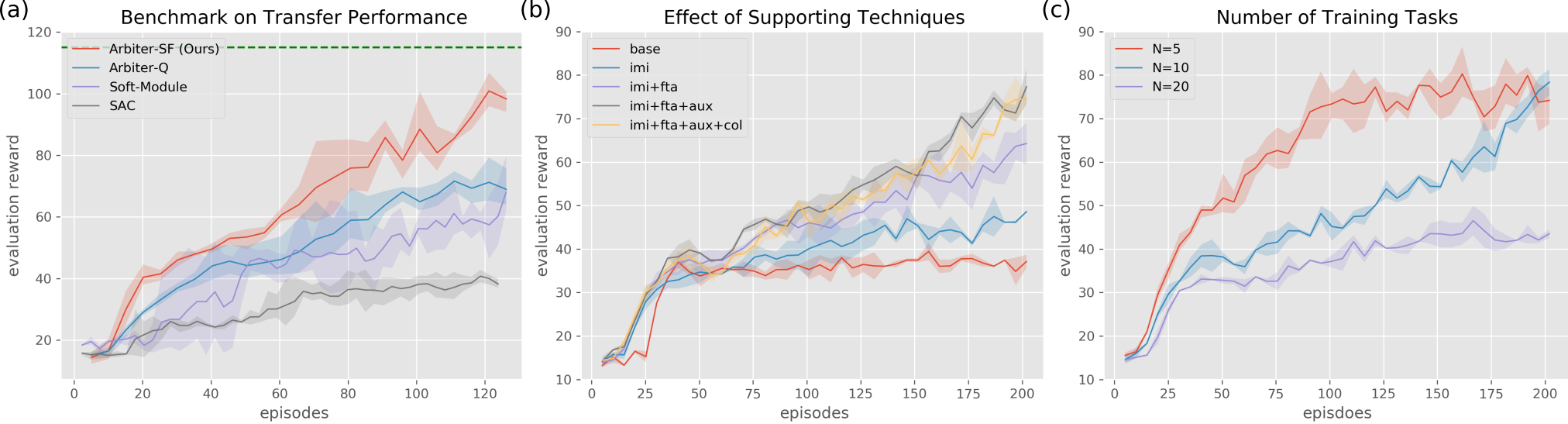}
		\phantomsubcaption
		\label{fig: transfer_performance}
	\end{subfigure}
	\begin{subfigure}[t]{0pt}
		\centering
		\phantomsubcaption
		\label{fig: ablation_supporting_techniques}
	\end{subfigure}
	\begin{subfigure}[t]{0pt}
		\centering
		\phantomsubcaption
		\label{fig: ablation_ntraintasks}
	\end{subfigure}
	\caption{\textbf{Experimental results on task transfer:} The solid line indicates the mean, and the top and bottom edges of the hue are the maximum and minimum rewards of each experiment. Compared to (a), the learning rate in experiments (b) and (c) is reduced ten times, but the number of episodes is increased from 125 to 200 to ensure training stability for better evaluation of the supporting techniques. Note that these experiments are conducted without domain randomization. Environment factors are set to default values and excluded from the agent's observation.
		(a) Benchmark on transfer performance in the evaluation task set. The green dotted line indicates the final performance of the agent deployed for real-world experiments. All agents incorporate imitation loss (Sec.~\ref{sec: Imitation Learning}). 
		(b) All the techniques, except collective learning, significantly improve the sample efficiency and transfer performance. \textit{base}: no techniques, \textit{imi}: imitation learning, \textit{fta}: fuzzy tiling activation, \textit{aux}: predictive auxiliary task, \textit{col}: collective learning. 
		(c) Training the Arbiter-SF with more tasks may improve final performance (115 for $N=10$ and 80 for $N=5$) at the cost of learning time. 
		\label{fig: Experimental results on task transfer}}
\end{figure*}

\subsection{Supporting Techniques}
\label{sec: Supporting Techniques}
We introduce compatible techniques for the adaptive agent to improve training stability and sample efficiency.

\subsubsection{Collective Learning}
\label{sec: Collective Learning}
An alternative to (\ref{eqn: soft successor feature policy improvement}) is to use the collective knowledge of multiple SFs. One can see all SFs, $\Psi^{\Pi}=\{\boldsymbol{\psi}^{\pi_{w_1}},...,\boldsymbol{\psi}^{\pi_{w_n}}  \}$, as an ensemble bundled to produce improved value estimation. In theory, the Q-value incurred from the ensemble, $ Q^{sup}_w = \max_{\boldsymbol{\psi}^{\pi} \in \Psi} \boldsymbol{\psi}^{\pi}\cdot w$,  is guaranteed to be greater or equal to any primitive Q-values in any task $w$ \cite{barreto2020fast}, i.e. $Q^{sup}_w \geq \sup_{\pi \in \Pi} Q^{\pi}_w$. That is, we can replace $Q^{\pi_{w_i}}_{w_i}$ with $Q^{sup}_{w_i}$ to conduct policy improvement step (\ref{eqn: soft successor feature policy improvement}) for $\pi_{w_i}$.  

\subsubsection{Imitation Learning}
\label{sec: Imitation Learning}
The arbiter architecture allows injecting prior knowledge to primitives via behavior cloning (BC) by minimizing the discrepancy between student and teacher actions:
\begin{align}
	J_{BC}(\theta_{\pi_{w_i}})=\mathbb{E}_{s\sim D}\mathbb{E}_{ \tiny \substack{a^{\pi_{w_i}} \sim \pi_{w_i} \\ a^{\pi_{tea}} \sim \pi_{tea}}} \left[ {||a^{\pi_{w_i}} - a^{\pi_{tea}}||_2} \right], \label{eqn: behavior clone}
\end{align}
where $\pi_{tea}$ is a teacher policy.
We need to assign primitives with their corresponding teachers and increment the BC loss to the respective primitive losses (\ref{eqn: primitive loss}), i.e. $J_{\pi_{w_i}}(\theta_{\pi_{w_i}})\mathrel{+}= k_{BC}\cdot J_{BC}(\theta_{\pi_{w_i}})$, where $k_{BC} \in \mathbb{R}^+$. This way, those primitives will replicate teacher actions and improve upon them based on their own primitive tasks. 

\subsubsection{Auxiliary Task}
\label{sec: Auxiliary Task}
The auxiliary task can improve the sample efficiency and representation by making auxiliary predictions \cite{jaderberg2016reinforcement} or solving auxiliary control task \cite{riedmiller2018learning}. 
In our case, one SFs head $j$, where $j>n$, is assigned to predict the next features for increasing the dynamics awareness by minimizing the objective, 
\begin{align}
	J_{Aux}(\theta_{\psi^{aux_j}})= \mathbb{E}_{\substack{s_t,a_t\sim D\\s_{t+1},a_{t+1}\sim D}}  \left[||\boldsymbol{\psi}^{aux_j}(s_t,a_t) - \boldsymbol{\phi}_{t+1}||_2 \right]. \label{eqn: auxtask}
\end{align}
where $k_{Aux} \in \mathbb{R}^+$. Additional auxiliary control tasks are introduced in Sec.~\ref{sec: Tasks}.

\subsubsection{Specialized Network Architecture}
\label{sec: Specialized Network Architecture}
The goal of network architecture (Fig.~\ref{fig: arbiter_architect}) is to develop a representation that can be shared among the primitives and avoid conflicting gradients \cite{yu2020gradient} to stabilize multitask learning. One way to reduce the conflict is by increasing the sparsity, for which we embed a fuzzy tiling activation function (FTA) before the final layer of primitive network \cite{pan2019fuzzy}.
We adopt residual network architecture \cite{sinha2020d2rl} and decoder $f^{dec}$ to construct the deeper network to preserve the information on environmental factors and enhance the agent's performance. The decoder loss is described as follows,
\begin{align}
	J_{f^{dec}}(\theta_{f^{dec}}, \theta_{f^{enc}}) =  ||e_t-f^{dec}(f^{enc}(e_t))||_2. \label{eqn: decoder loss}
\end{align}
It is incremented with SFs loss (\ref{eqn: sfloss}) to update the encoder.


\section{Experiments and Results}
\label{sec:4_experiment}
The experiments aim to answer the questions: (1) How well is the adaptive agent solving unseen tasks? (2) What are the critical aspects of successfully training an adaptive agent? 

\subsection{Blimp Control Problem}

\subsubsection{State, Action, and Feature Space} 
The blimp control's state, action, and feature space are described in Table.~\ref{tab:env_space}. At every time step, each primitive observes the state of the robot, goal, and environment and produces an action to control the blimp. 
\begin{table}[h]
	\centering
	\resizebox{0.4\textwidth}{!}{%
		\begin{tabular}{ c||c }
			\hline
			$\mathcal{S}$ & $pose^{robot}, pose^{goal_{nav}}, pose^{goal_{hov}}, s^{env}$   \\
			\hline
			$\mathcal{A}$ & $a^{thrust}, a^{servo}, a^{pitch}, a^{yaw}$  \\
			\hline
			
	\end{tabular}}
	\caption{State and action space for the blimp control \label{tab:env_space} 
	}
\end{table}
The state space is constituted of the robot, goal, and environment states, where $pose=(\mathbb{X}, \mathbb{V}, \Theta, \Omega)\in(\mathbb{R}^3)^4$ denotes the position, velocity, angle, and angular velocity of the robot and goals. For convenience, we define two goals, a navigation and a hover goal (Fig.~\ref{fig: illustration}). The hover goal $pose^{goal_{hov}}$ is placed at the center of the flight zone, whereas the navigation goals $pose^{goal_{nav}}$ are waypoints defined within the flight zone. 
The extracted environment state $S^{env} \in \mathbb{R}^{15}$ has an entry length defined by the user. 
In practice, actuator states are included, e.g., thrust and servo angle.
The action space is a vector of four entries $\mathcal{A}\in[-1, 1]^4$ (Table.~\ref{tab:env_space}), where $a^{thrust}$ controls two motors, $a^{servo}$ controls the thrusting angle, $a^{pitch}$ controls the left and the right elevators, and $a^{yaw}$ controls the top and bottom rudders and a bottom motor. We define the task-relevant features $\boldsymbol{\phi}\in\mathbb{R}^{11}$ as follows (\ref{eqn: features computation}),
\begin{gather}
	\footnotesize
	\begin{split}
		&\boldsymbol{\phi} = \\
		&\begin{bmatrix} 
			\phi^{dist_{xy}} \\
			\phi^{dist_z} \\
			\phi^{trigger} \\    
			\phi^{yaw_{heading}} \\
			\phi^{proximity} \\
			\phi^{yaw} \\  
			\phi^{v_{heading}} \\  
			\phi^{v_{xy}} \\
			\phi^{v_{z}} \\  
			\phi^{reg_{RP}} \\  
			\phi^{reg_T} 
		\end{bmatrix}
		=\begin{bmatrix}
			||\mathbb{X}_{xy}^{robot}-\mathbb{X}_{xy}^{goal_{nav}}||_{\mathcal{N}} \\ 
			||\mathbb{X}_{z}^{robot}-\mathbb{X}_{z}^{goal_{nav}}||_{\mathcal{N}} \\ 
			1 ~\text{if}~ ||\mathbb{X}^{robot}-\mathbb{X}^{goal_{nav}}||_{2} <5 ~\text{else}~ 0 \\
			||\Theta_{z_{\mathcal{B}}}^{robot}-\Theta_{z}^{goal_{nav}}||_{\mathcal{N}} \\ 
			1 ~\text{if}~ ||\mathbb{X}^{robot}-\mathbb{X}^{goal_{hov}}||_{2} <7 ~\text{else}~ 0 \\ 		
			||\Theta_z^{robot}-\Theta_z^{goal_{hov}}||_{\mathcal{N}} \\ 
			||v_{x_{\mathcal{B}}}^{robot}-|v^{goal_{hov}}|||_{\mathcal{N}} \\ 
			||v_{xy}^{robot}-v_{xy}^{goal_{nav}}||_{\mathcal{N}} \\ 
			||v_{z}^{robot}-v_{z}^{goal_{nav}}||_{\mathcal{N}} \\ 
			||\Theta_{xy}^{robot}||_{\mathcal{N}} \\ 
			||a_{thurst}+1||_{2} 
		\end{bmatrix} \label{eqn: features computation}
	\end{split} 
\end{gather} 
where $||\cdot||_2$ denotes L2 norm, the Gaussian norm is defined as $||\cdot||_{\mathcal{N}}\equiv e^{-k ||\cdot||_2 / \sigma^2}$
and subscript ${\mathcal{B}}$ denotes state in body frame.
These features measure the differences between robot and goal states for defining tasks. The first four are relevant to position control, the next three to hover control, the eighth and ninth to velocity control, and the last to regularization tasks. 

\begin{figure*}
	\centering
	\begin{minipage}[t]{0.32\textwidth}
		\centering
		\includegraphics[width=0.95\textwidth]{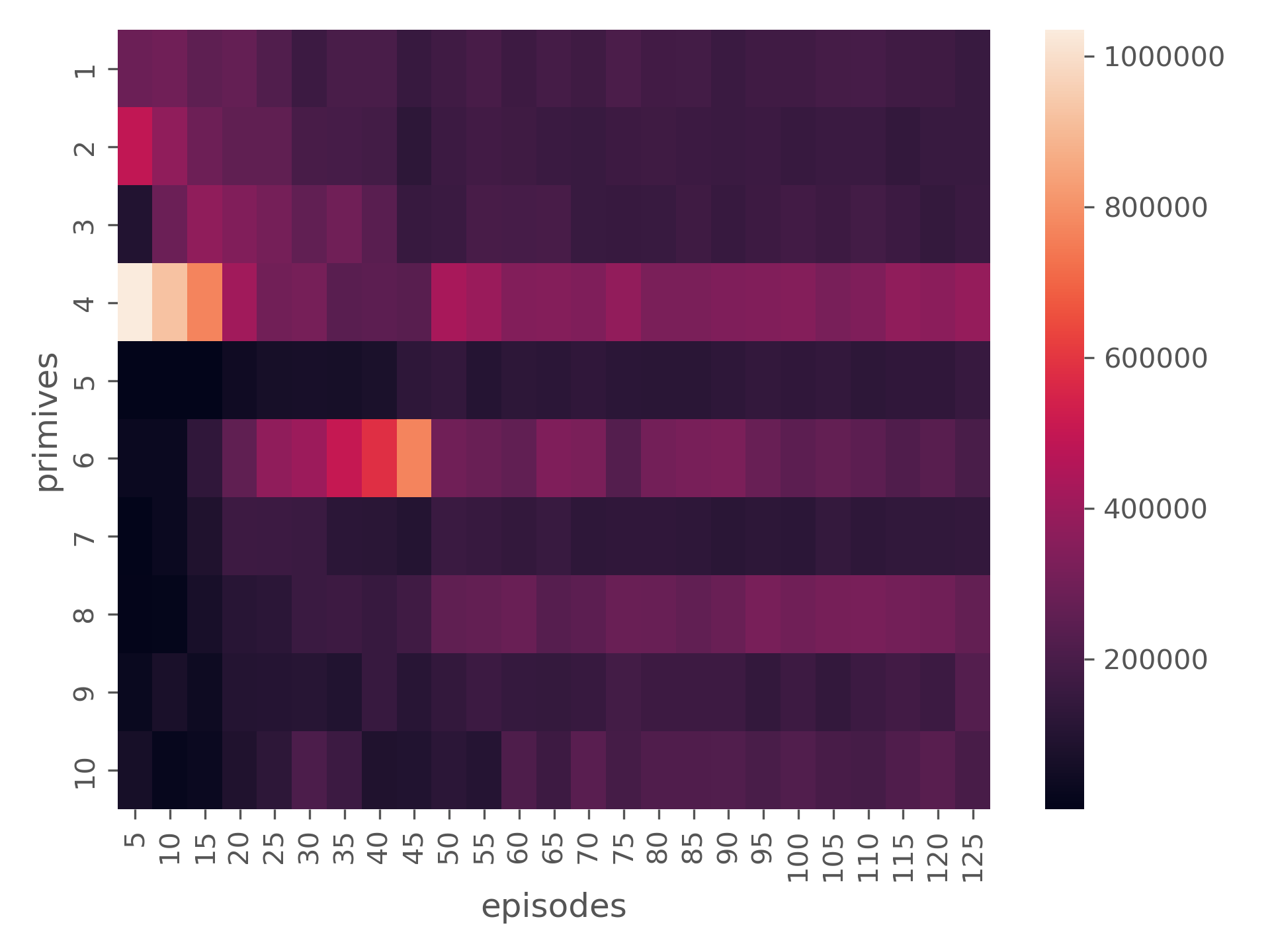}
		\caption{The color coding indicates the counts of each primitive being chosen to perform the action by the arbiter in an evaluation episode every five training episodes.} 
		\label{fig: activation_frequency}
	\end{minipage}
	\hfill
	\begin{minipage}[t]{0.67\textwidth}
		\centering
		\captionsetup[subfigure]{labelformat=empty}
		\begin{subfigure}[t]{0.95\textwidth}
			\centering
			\includegraphics[width=0.99\textwidth]{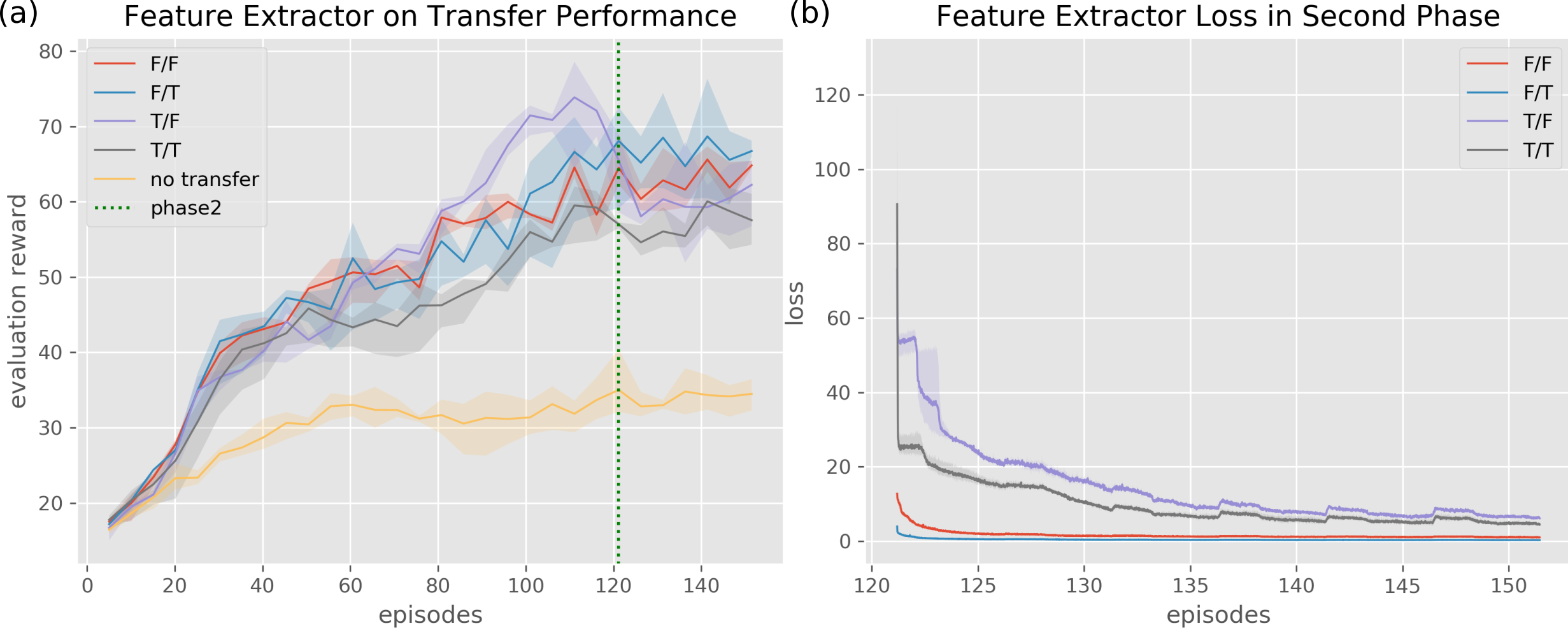}
			\phantomsubcaption
			\label{fig: domain_rand}
		\end{subfigure}
		\hfill
		\begin{subfigure}[t]{0\textwidth}
			\centering
			\phantomsubcaption
			\label{fig: extractor_loss}
		\end{subfigure}
		\caption{\textbf{Experiment results on domain transfer:} The label indicates if the decoder/residual network is applied, and T or F denotes True or False. The experiments apply domain randomization and utilize the best configuration from \ref{fig: ablation_supporting_techniques} with an encoder and extractor incorporated. (a) The yellow line indicates learning with only the extractor. All configures drop in performance when entering the second phase (Green). (b) The extractor loss (\ref{eqn: extractor loss}) is higher with the decoder but reduced with the residual network.
		}
	\end{minipage}
\end{figure*}

\subsubsection{Tasks} 
\label{sec: Tasks}
We have in total ten primitive tasks as the training task set $\boldsymbol{w}=\{w_1,...,w_{10}\}$ and each has the same dimension as the features, i.e., $w_i \in \mathbb{R}^{11}$, defined in (\ref{eqn: blimp primitive tasks}).
The first three tasks of interest are position control, hover control, and velocity control, denoted by $w_1$, $w_2$, and $w_3$, respectively. In position control, the objective is to sequentially trigger waypoints by reducing the distance to the current one to be less than five meters. Then, the next waypoint will become activated. The hover task requires the blimp to maintain its position within a specified range for the longest possible duration. Velocity control, similar to position control, involves activating waypoints, but it is guided by velocity commands rather than positional ones. A planner provides velocity commands based on the relative position to navigation goal at every time step, i.e.  $\mathbb{V}^{goal_{nav}}=6\cdot (\mathbb{X}^{goal_{nav}}-\mathbb{X}^{robot})/||\mathbb{X}^{goal_{nav}}-\mathbb{X}^{robot}||_1$. The remaining tasks $w_4-w_{10}$ serve as auxiliary control tasks (as mentioned in Sec.~\ref{sec: Auxiliary Task}), offering alternative methods for tackling similar tasks or enhancing specific abilities, e.g., $w_8$ focuses solely on yaw angle control.

\begin{gather}
	\footnotesize
	\boldsymbol{w} = 
	\begin{bmatrix} 
		w_1 \\
		w_2 \\
		w_3 \\
		w_4 \\
		w_5 \\
		w_6 \\
		w_7 \\
		w_8 \\
		w_9 \\
		w_{10} 
	\end{bmatrix}
	=\begin{bmatrix}
		.1,.1,1,0,&  0,0,0,    &0,0,  &.01,0 \\
		0,0,0,0,&    10,0,.01, &0,0,  &0,.1 \\
		0,0,0,0,&    0,0,0,    &1,.5, &.01,0 \\
		.1,.1,0,-1,& 0,0,-1,   &0,0,  &0,0 \\
		.1,.5,.1,1,& 0,0,1,    &0,0,  &.01,.01 \\
		0,0,0,0,&    10,.2,0,  &0,0,  &0,.5 \\
		0,0,.1,.1,&  0,0,0,    &1,.5, &.01,0 \\
		0,0,0,0,&    0,1,0,    &0,0,  &0,.1 \\ 
		1,.1,0,.1,&  0,0,0,    &0,0,  &0,0\\
		0,0,0,0,&    1,0,0,    &0,0,  &.1,.1
	\end{bmatrix}. \label{eqn: blimp primitive tasks}
\end{gather}

To accelerate training, knowledge from PID control-based teachers is incorporated through behavior cloning (Sec.~\ref{sec: Imitation Learning}) to assist in solving the first four tasks $w_1-w_4$. Given that PID teachers are optimized for nominal environment conditions and perform sub-optimally outside these conditions, blindly imitating them can lead to poor performance for the primitives.  

\begin{gather}
	\footnotesize
	\boldsymbol{w}^{eval} = 
	\begin{bmatrix} 
		w^{eval}_1 \\
		w^{eval}_2 \\
		w^{eval}_3 \\
		w^{eval}_4 \\
		w^{eval}_5 \\
		w^{eval}_6 \\
		w^{eval}_7 
	\end{bmatrix}
	=\begin{bmatrix}
		1,0,0,0,& 0,0,0, &0,0, &0,0\\
		0,0,1,0,& 0,0,0, &0,0, &0,0\\
		0,0,0,1,& 0,0,1, &0,0, &0,0\\
		0,0,0,0,& 1,0,0, &0,0, &0,0\\
		0,0,0,0,& 1,1,0, &0,0, &0,0\\
		0,0,0,0,& 0,0,0, &1,0, &0,0\\
		0,0,0,0,& 0,0,0, &1,.5, &0,0
	\end{bmatrix}. \label{eqn: blimp evaluation tasks}
\end{gather}

For evaluation, we define a set of evaluation tasks $\mathbf{w}^{eval}$ (\ref{eqn: blimp evaluation tasks}), different from the training set, to test the agent's transfer ability. These tasks also cover basic tasks, including position, velocity, and hover control, with some arbitrarily defined new unseen tasks. 
For each of the parallelized simulations, we sample a task uniform randomly from the task set and apply an arbiter to perform task transfer. 

\begin{figure*}
	\centering
	\captionsetup[subfigure]{labelformat=empty}
	\begin{subfigure}[t]{0.87\textwidth}
		\centering
		\includegraphics[width=0.99\textwidth]{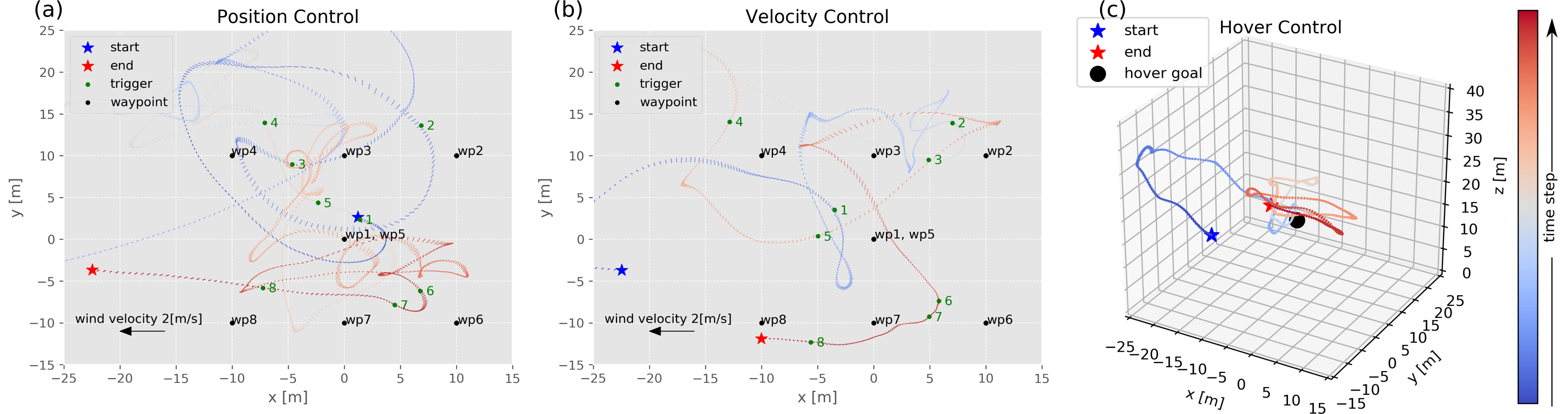}
		\phantomsubcaption
		\label{fig:realtasks_a}
	\end{subfigure}
	\begin{subfigure}[t]{0pt}
		\centering
		\phantomsubcaption
		\label{fig:realtasks_b}
	\end{subfigure}
	\begin{subfigure}[t]{0pt}
		\centering
		\phantomsubcaption
		\label{fig:realtasks_c}
	\end{subfigure}
	\caption{\textbf{Real world experiments:} robot trajectories for different tasks performed in the real-world experiment. For visual clarity, only tasks involving changes in altitude are depicted in 3-D format. (a) The agent triggers all the waypoints by following the position command. $w=[.5,.2,1,.1, 0,0,0, 0,0, .1,0]$. (b) The agent triggers all the waypoints by following the velocity command. $w=[0,0,1,0, 0,0,0, 1,.5, .1,0]$. (c) Hover tasks bring the blimp near the hover region at the center of flight zone $w=[0,0,0,0, 1,0,0, 0,0, 0,1]$.
		\label{fig:realtasks}}
\end{figure*}

\subsection{Experiment Setup}
\label{sec: Experiment Setup}
We build our simulation based on IsaacGym \cite{liang2018gpu} to facilitate data collection in parallel and enable multi-task learning. The adaptive agent is built on Pytorch with just-in-time compilation to optimize the code performance. 

For the blimp control, the first RMA training phase is set to $1000$ episodes (13.9 hrs) and $250$ episodes (3.8 hrs) for the second phase, with both using the training task set (\ref{eqn: blimp primitive tasks}). Each episode contains 500 environment steps with 1024 environments running in parallel, with each environment step set to 0.1 seconds. The number of episodes is reduced in the ablation study to save the amount of computation.
In every episode, domain randomization is applied to generate goals with different poses and environmental factors, as shown in Table.~\ref{tab: env latents}. To ensure the motion smoothness, instead of directly mapping the agent action to the thrust, a motor smoothing factor is applied via moving average with previous actions, i.e., $thrust_t=c_1\cdot(c_2\cdot a^{thrust}_t+(1-c_2)\cdot a^{thrust}_{t-1})$, where $(c_1, c_2)=(10,0.2)$. 
\begin{table}[h]
	\centering
	\resizebox{0.4\textwidth}{!}{%
		\begin{tabular}{ c||c }
			\hline
			Type A & body area, weight distribution, \\ 
			& wind magnitude and variance, blimp mass, \\ 
			& lift/drag coefficients, buoyancy \\
			\hline
			Type B & thrust strength $(c_1)$, motor smoothing factor $(c_2)$ \\
			\hline
	\end{tabular}}
	\caption{Environment Factors. Type A variables are randomized in the range of 80-125\% from the nominal condition, whereas 50-200\% for type B variables. Note that the mean is set to the mass and drawn from the uniform distribution to keep buoyancy close to the mass. \label{tab: env latents} 
	}
\end{table}

The real-world experiment is performed on a five-meter blimp (Fig.~\ref{fig: illustration}), equipped with a GPS sensor and a Librepilot onboard flight controller, which has a built-in IMU and a gyroscope sensor. The state estimation is then obtained via an extended Kalman filter. 
The adaptive agent is mounted on the onboard GPU, Nvidia TX2, and runs at 10Hz. The task is assigned by the user on the laptop and transmitted to the agent via ROS multi-master service in \textit{real-time}.

\subsection{Experiment Result}


\subsubsection{\textbf{Task Transfer in Simulated Environment}}
In nominal environment conditions, our agent can perform zero-shot transfer to arbitrary tasks, as depicted in Fig.~\ref{fig:simtasks}. The hover and angle control task (Fig.~\ref{fig:simtasks_a}) positions the blimp in the hover region with the desired angle. The task is useful for tracking with a fixed camera \cite{price2023driven}. By increasing the weight of thrust regularization, we can prevent overshooting and reduce energy consumption. Integrating hover and velocity control task (Fig.~\ref{fig:simtasks_b}) allows to track the desired velocity within hover region. Sometimes, it is more efficient to stay in motion by leveraging aerodynamics over constant vertical thrusting. The position and velocity control (Fig.~\ref{fig:simtasks_c}) can be combined to improve waypoint tracking. Blindly following the velocity command often leads to overshooting, like the velocity PID teacher, and including position control can mitigate this.  Lastly, the backward position control (Fig.~\ref{fig:simtasks_d}) cannot be achieved. The agent learns to point the blimp's tail to the waypoint without maintaining a constant velocity. Still, backward control is practical for reducing velocity or when only a tiny correction from overshooting is required.

\subsubsection{\textbf{Asymptotic Performance and Task Transfer-ability}} 
We benchmark our adaptive agent -- SF-based arbiter (Arbiter-SF) -- against other SAC-based agents for the blimp control task on asymptotic task transfer performance:
\begin{itemize}
	\item SAC \cite{haarnoja2018soft}: A state-of-the-art single task agent. Task weight is augmented to the observation for both Q-value and policy networks, i.e., $s\leftarrow (s, w)$.
	\item Arbiter-Q: A Q-function-based arbiter derived from SAC. Task weight is augmented to the observation for both Q-value and primitive networks. 
	\item Soft-Module \cite{yang2020multi}: A state-of-the-art multi-task agent with task embedding replaced by the task weight. 
\end{itemize}
All agents are trained with the task set $\boldsymbol{w}$ (\ref{eqn: blimp primitive tasks}) and evaluated with $\boldsymbol{w}^{eval}$ (\ref{eqn: blimp evaluation tasks}). To ensure a fair comparison, all supporting techniques were excluded from the Arbiter-SF, except for the imitation loss, which was incorporated into all agents, i.e., Arbiter-Q, SAC, and Soft-Module, to promote quicker convergence. Following a hyper-parameter tuning by the Bayesian optimization 25 times, the top performance for each agent was selected for comparison. For the blimp control, our Arbiter-SF achieves the best transfer performance after 125 episodes of training (Fig.~\ref{fig: transfer_performance}). 



\subsubsection{\textbf{Ablation Study on Task Transfer}} 
\label{sec: Ablation Study on Task Transfer}
All techniques (Sec.~\ref{sec: Supporting Techniques}) contribute significantly except collective learning (Fig.~\ref{fig: ablation_supporting_techniques}), implying that the estimated Q-values are nearly identical by using the primitive's SFs or by the collective SFs. Next, Fig.~\ref{fig: ablation_ntraintasks} shows that there is a trade-off between the number of primitives on learning speed and final performance, contrary to the idea that incorporating auxiliary tasks should increase sample efficiency \cite{riedmiller2018learning, jaderberg2016reinforcement}.

\subsubsection{\textbf{Primitive Evaluation}}
The learning pace of the primitives varies significantly, as shown in Fig.~\ref{fig: activation_frequency}. At the start of training, the agent predominantly activates the first four primitives, which imitate PID teachers. Notably, the fourth primitive, guided by a backward flight PID, has the highest activation frequency. This may indicate that one initial learning outcome is decelerating the blimp's speed. As training progresses, the activation frequency of primitives becomes more balanced, indicating that most primitives have developed skills in task-solving. 



\subsubsection{\textbf{Ablation Study on Domain Transfer}} 
Fig.~\ref{fig: domain_rand} shows that applying residual networks for encoder and extractor improves transfer performance, but adding a decoder decreases it. Further examination in the second phase (Fig.~\ref{fig: extractor_loss}) reveals that the presence of a decoder results in significantly higher loss, suggesting that the resulting representation is more complex and potentially richer in content. However, this increased richness does not translate to more effective task-solving content with agents trained in limited episodes. Still, we recommend incorporating a decoder into the architecture as it can benefit agents trained over longer episodes.

\subsubsection{\textbf{Real World Experiment}}
\label{sec: Real World Experiment}
\begin{figure}[h]
	\centering
	\includegraphics[width=0.49\textwidth]{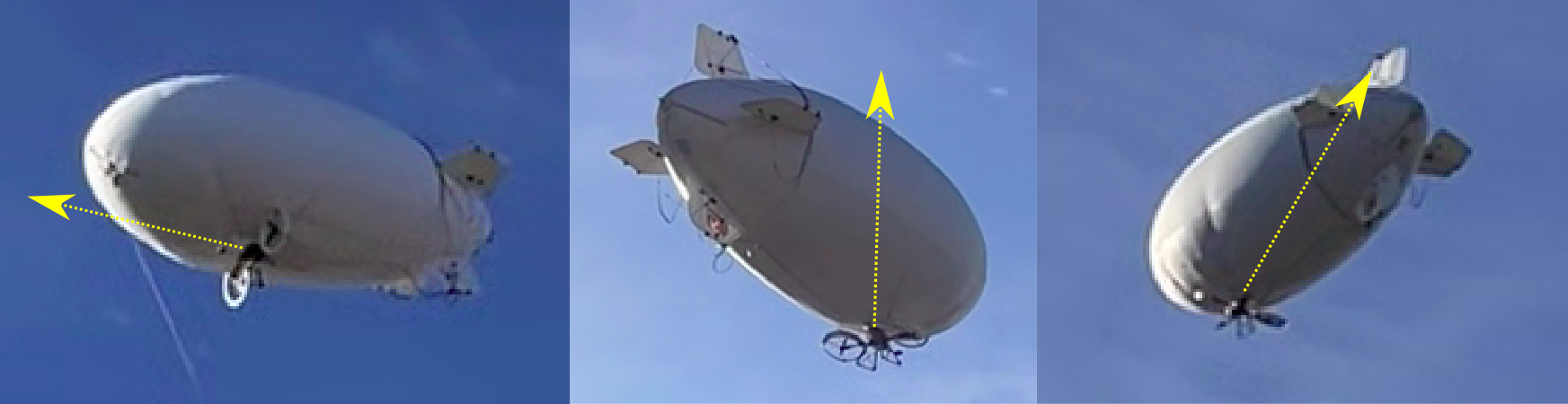}	
	\caption{Notable behaviors include extensive use of the thrust vector (yellow arrow), such as forward (left), upward (middle), and backward thrust (right).}
	\label{fig: emerging_behavior}
\end{figure}

The trajectory of the blimp is displayed in Fig.~\ref{fig:realtasks}. Initially, a human pilot manually takes the blimp off to around 7 meters for safety purposes, after which the agent takes over the control to complete the ascent to 20 meters within the designated hover region. Then, the tasks, including position, velocity, and hover control, are executed sequentially. In each starting phase of the task, the human operator needs to tune the task weight. 
During the experiment, we observed a different flight pattern from the classic PID controller (Fig.~\ref{fig: emerging_behavior}). PID controllers generally rely on forward or upward thrust to maneuver the blimp, struggling with thrust angles between or beyond ninety degrees due to non-linearity and the dynamic inversion problem. In contrast, the agent extensively employs thrust vectoring, achieving a more agile flight style. We refer more discussion on DRL and PID methods to \cite{liu2022deep}, where our PID teachers were benchmarked.

\section{Conclusions and Future Work}
\label{sec:5_discussion}
In this work, we proposed an adaptive agent that supports task and domain transfer utilizing the arbiter architecture with value estimation provided by the successor feature. The experiments on the blimp control tasks demonstrate that our agent achieves the best transfer performance on unseen tasks and enables zero-shot transfer to different tasks and domains, validated in real-world experiments.
One drawback, as mentioned in Sec.~\ref{sec: Ablation Study on Task Transfer}, is that providing more auxiliary control tasks did not always improve sample efficiency. Furthermore, sometimes, we observe action inconsistency, which implies that the SFs' value estimation can be noisy and disturb the arbiter's decision-making. Our future work will focus on increasing the action consistency and examining the side effects accompanied by the auxiliary control tasks.


	\bibliographystyle{IEEEtran}
	\bibliography{biblio}

\begin{thebibliography}{10}
\providecommand{\url}[1]{#1}
\csname url@rmstyle\endcsname
\providecommand{\newblock}{\relax}
\providecommand{\bibinfo}[2]{#2}
\providecommand\BIBentrySTDinterwordspacing{\spaceskip=0pt\relax}
\providecommand\BIBentryALTinterwordstretchfactor{4}
\providecommand\BIBentryALTinterwordspacing{\spaceskip=\fontdimen2\font plus
\BIBentryALTinterwordstretchfactor\fontdimen3\font minus
  \fontdimen4\font\relax}
\providecommand\BIBforeignlanguage[2]{{%
\expandafter\ifx\csname l@#1\endcsname\relax
\typeout{** WARNING: IEEEtran.bst: No hyphenation pattern has been}%
\typeout{** loaded for the language `#1'. Using the pattern for}%
\typeout{** the default language instead.}%
\else
\language=\csname l@#1\endcsname
\fi
#2}}

\bibitem{bertsekas1996neuro}
D.~Bertsekas and J.~N. Tsitsiklis, \emph{Neuro-dynamic programming}.\hskip 1em
  plus 0.5em minus 0.4em\relax Athena Scientific, 1996.

\bibitem{zhu2023transfer}
Z.~Zhu, K.~Lin, A.~K. Jain, and J.~Zhou, ``Transfer learning in deep
  reinforcement learning: A survey,'' \emph{IEEE Transactions on Pattern
  Analysis and Machine Intelligence}, 2023.

\bibitem{zhao2020sim}
W.~Zhao, J.~P. Queralta, and T.~Westerlund, ``Sim-to-real transfer in deep
  reinforcement learning for robotics: a survey,'' in \emph{2020 IEEE symposium
  series on computational intelligence (SSCI)}.\hskip 1em plus 0.5em minus
  0.4em\relax IEEE, 2020, pp. 737--744.

\bibitem{russell2003q}
S.~J. Russell and A.~Zimdars, ``Q-decomposition for reinforcement learning
  agents,'' in \emph{Proceedings of the 20th International Conference on
  Machine Learning (ICML-03)}, 2003, pp. 656--663.

\bibitem{barreto2020fast}
A.~Barreto, S.~Hou, D.~Borsa, D.~Silver, and D.~Precup, ``Fast reinforcement
  learning with generalized policy updates,'' \emph{Proceedings of the National
  Academy of Sciences}, vol. 117, no.~48, pp. 30\,079--30\,087, 2020.

\bibitem{kumar2021rma}
A.~Kumar, Z.~Fu, D.~Pathak, and J.~Malik, ``Rma: Rapid motor adaptation for
  legged robots,'' \emph{arXiv preprint arXiv:2107.04034}, 2021.

\bibitem{price2023driven}
E.~Price, M.~J. Black, and A.~Ahmad, ``Viewpoint-driven formation control of
  airships for cooperative target tracking,'' \emph{IEEE Robotics and
  Automation Letters}, 2023.

\bibitem{liu2022deep}
Y.~T. Liu, E.~Price, M.~J. Black, and A.~Ahmad, ``Deep residual reinforcement
  learning based autonomous blimp control,'' in \emph{2022 IEEE/RSJ
  International Conference on Intelligent Robots and Systems (IROS)}.\hskip 1em
  plus 0.5em minus 0.4em\relax IEEE, 2022, pp. 12\,566--12\,573.

\bibitem{liang2018gpu}
J.~Liang, V.~Makoviychuk, A.~Handa, N.~Chentanez, M.~Macklin, and D.~Fox,
  ``Gpu-accelerated robotic simulation for distributed reinforcement
  learning,'' in \emph{Conference on Robot Learning}.\hskip 1em plus 0.5em
  minus 0.4em\relax PMLR, 2018, pp. 270--282.

\bibitem{espeholt2018impala}
L.~Espeholt, H.~Soyer, R.~Munos, K.~Simonyan, V.~Mnih, T.~Ward, Y.~Doron,
  V.~Firoiu, T.~Harley, I.~Dunning, \emph{et~al.}, ``Impala: Scalable
  distributed deep-rl with importance weighted actor-learner architectures,''
  in \emph{International conference on machine learning}.\hskip 1em plus 0.5em
  minus 0.4em\relax PMLR, 2018, pp. 1407--1416.

\bibitem{kalashnikov2021mt}
D.~Kalashnikov, J.~Varley, Y.~Chebotar, B.~Swanson, R.~Jonschkowski, C.~Finn,
  S.~Levine, and K.~Hausman, ``Mt-opt: Continuous multi-task robotic
  reinforcement learning at scale,'' \emph{arXiv preprint arXiv:2104.08212},
  2021.

\bibitem{peng2019mcp}
X.~B. Peng, M.~Chang, G.~Zhang, P.~Abbeel, and S.~Levine, ``Mcp: Learning
  composable hierarchical control with multiplicative compositional policies,''
  \emph{Advances in Neural Information Processing Systems}, vol.~32, 2019.

\bibitem{pateria2021hierarchical}
S.~Pateria, B.~Subagdja, A.-h. Tan, and C.~Quek, ``Hierarchical reinforcement
  learning: A comprehensive survey,'' \emph{ACM Computing Surveys (CSUR)},
  vol.~54, no.~5, pp. 1--35, 2021.

\bibitem{sutton1999between}
R.~S. Sutton, D.~Precup, and S.~Singh, ``Between mdps and semi-mdps: A
  framework for temporal abstraction in reinforcement learning,''
  \emph{Artificial intelligence}, vol. 112, no. 1-2, pp. 181--211, 1999.

\bibitem{yang2020multi}
R.~Yang, H.~Xu, Y.~Wu, and X.~Wang, ``Multi-task reinforcement learning with
  soft modularization,'' \emph{Advances in Neural Information Processing
  Systems}, vol.~33, pp. 4767--4777, 2020.

\bibitem{ghosh2017divide}
D.~Ghosh, A.~Singh, A.~Rajeswaran, V.~Kumar, and S.~Levine,
  ``Divide-and-conquer reinforcement learning,'' \emph{arXiv preprint
  arXiv:1711.09874}, 2017.

\bibitem{gupta2017learning}
A.~Gupta, C.~Devin, Y.~Liu, P.~Abbeel, and S.~Levine, ``Learning invariant
  feature spaces to transfer skills with reinforcement learning,'' \emph{arXiv
  preprint arXiv:1703.02949}, 2017.

\bibitem{zhang2018decoupling}
A.~Zhang, H.~Satija, and J.~Pineau, ``Decoupling dynamics and reward for
  transfer learning,'' \emph{arXiv preprint arXiv:1804.10689}, 2018.

\bibitem{dayan1993improving}
P.~Dayan, ``Improving generalization for temporal difference learning: The
  successor representation,'' \emph{Neural computation}, vol.~5, no.~4, pp.
  613--624, 1993.

\bibitem{liu2024multi}
Y.~T. Liu and A.~Ahmad, ``Multi-task reinforcement learning in continuous
  control with successor feature-based concurrent composition,'' in \emph{2024
  European Control Conference (ECC)}.\hskip 1em plus 0.5em minus 0.4em\relax
  IEEE, 2024, pp. 3860--3867.

\bibitem{tobin2017domain}
J.~Tobin, R.~Fong, A.~Ray, J.~Schneider, W.~Zaremba, and P.~Abbeel, ``Domain
  randomization for transferring deep neural networks from simulation to the
  real world,'' in \emph{2017 IEEE/RSJ international conference on intelligent
  robots and systems (IROS)}.\hskip 1em plus 0.5em minus 0.4em\relax IEEE,
  2017, pp. 23--30.

\bibitem{taylor2009transfer}
M.~E. Taylor and P.~Stone, ``Transfer learning for reinforcement learning
  domains: A survey.'' \emph{Journal of Machine Learning Research}, vol.~10,
  no.~7, 2009.

\bibitem{ho2016generative}
J.~Ho and S.~Ermon, ``Generative adversarial imitation learning,''
  \emph{Advances in neural information processing systems}, vol.~29, 2016.

\bibitem{wang2016learning}
J.~X. Wang, Z.~Kurth-Nelson, D.~Tirumala, H.~Soyer, J.~Z. Leibo, R.~Munos,
  C.~Blundell, D.~Kumaran, and M.~Botvinick, ``Learning to reinforcement
  learn,'' \emph{arXiv preprint arXiv:1611.05763}, 2016.

\bibitem{peng2020learning}
X.~B. Peng, E.~Coumans, T.~Zhang, T.-W. Lee, J.~Tan, and S.~Levine, ``Learning
  agile robotic locomotion skills by imitating animals,'' \emph{arXiv preprint
  arXiv:2004.00784}, 2020.

\bibitem{price2020simulation}
E.~Price, Y.~T. Liu, M.~J. Black, and A.~Ahmad, ``Simulation and control of
  deformable autonomous airships in turbulent wind,'' in \emph{16th
  International Conference on Intelligent Autonomous System (IAS)}, June 2021.

\bibitem{rottmann2009adaptive}
A.~Rottmann and W.~Burgard, ``Adaptive autonomous control using online value
  iteration with gaussian processes,'' in \emph{2009 IEEE International
  Conference on Robotics and Automation}.\hskip 1em plus 0.5em minus
  0.4em\relax IEEE, 2009, pp. 2106--2111.

\bibitem{jaderberg2016reinforcement}
M.~Jaderberg, V.~Mnih, W.~M. Czarnecki, T.~Schaul, J.~Z. Leibo, D.~Silver, and
  K.~Kavukcuoglu, ``Reinforcement learning with unsupervised auxiliary tasks,''
  2016.

\bibitem{riedmiller2018learning}
M.~Riedmiller, R.~Hafner, T.~Lampe, M.~Neunert, J.~Degrave, T.~Wiele, V.~Mnih,
  N.~Heess, and J.~T. Springenberg, ``Learning by playing solving sparse reward
  tasks from scratch,'' in \emph{International conference on machine
  learning}.\hskip 1em plus 0.5em minus 0.4em\relax PMLR, 2018, pp. 4344--4353.

\bibitem{yu2020gradient}
T.~Yu, S.~Kumar, A.~Gupta, S.~Levine, K.~Hausman, and C.~Finn, ``Gradient
  surgery for multi-task learning,'' \emph{Advances in Neural Information
  Processing Systems}, vol.~33, pp. 5824--5836, 2020.

\bibitem{pan2019fuzzy}
Y.~Pan, K.~Banman, and M.~White, ``Fuzzy tiling activations: A simple approach
  to learning sparse representations online,'' \emph{arXiv preprint
  arXiv:1911.08068}, 2019.

\bibitem{sinha2020d2rl}
S.~Sinha, H.~Bharadhwaj, A.~Srinivas, and A.~Garg, ``D2rl: Deep dense
  architectures in reinforcement learning,'' \emph{arXiv preprint
  arXiv:2010.09163}, 2020.

\bibitem{haarnoja2018soft}
T.~Haarnoja, A.~Zhou, P.~Abbeel, and S.~Levine, ``Soft actor-critic: Off-policy
  maximum entropy deep reinforcement learning with a stochastic actor,'' in
  \emph{International conference on machine learning}.\hskip 1em plus 0.5em
  minus 0.4em\relax PMLR, 2018, pp. 1861--1870.

\end{thebibliography}
\end{document}